\documentclass{article} 
\usepackage[preprint]{neurips_2025}

\usepackage{amsmath,amsfonts,bm}

\def\eqref#1{equation~\ref{#1}}

\def\1{\bm{1}}

\DeclareMathAlphabet{\mathsfit}{\encodingdefault}{\sfdefault}{m}{sl}
\SetMathAlphabet{\mathsfit}{bold}{\encodingdefault}{\sfdefault}{bx}{n}

\usepackage{hyperref}
\usepackage{url}
\usepackage{graphicx}
\usepackage{subfigure}
\usepackage[capitalize,noabbrev]{cleveref}
\usepackage{booktabs}
\usepackage{multirow}
\usepackage{tcolorbox}
\usepackage[table]{xcolor}
\usepackage{caption}

\title{Information-Preserving Reformulation of Reasoning Traces for Antidistillation}

\author{
\vspace{-0.25in} \\
\textbf{
Jiayu Ding$^{\spadesuit}$~~~~Lei Cui$^{\clubsuit}$~~~Li Dong$^{\clubsuit}$~~~{Nanning Zheng}$^{\spadesuit}$\footnotemark[2]~~~{Furu Wei}$^{\clubsuit}$\footnotemark[2]} \\[3pt]
$^{\spadesuit}$IAIR, Xi'an Jiaotong University \\[3pt]
$^{\clubsuit}$Microsoft Research \\
\vspace{-0.4cm}
\\}

\newcommand\ours{\textsc{PART}}
\renewcommand{\thefootnote}{\fnsymbol{footnote}}

\begin{document}

\maketitle

\footnotetext[2]{Corresponding author.}
\renewcommand{\thefootnote}{\arabic{footnote}}

\begin{abstract}
Recent advances in Large Language Models (LLMs) show that extending the length of reasoning chains significantly improves performance on complex tasks. While revealing these reasoning traces helps users better follow, verify, and learn from the model’s problem-solving process, it also makes them highly vulnerable to unauthorized distillation. To mitigate this risk, proprietary model providers often adopt aggressive protection strategies, such as replacing detailed reasoning with brief summaries, which deprive users of valuable intermediate information. 
To address this trade-off, we propose \ours{}, an information-\textbf{P}reserving \textbf{A}ntidistillation \textbf{R}eformulation of reasoning \textbf{T}races. Motivated by the difference between how humans understand reasoning traces and how LLMs exploit them for supervised fine-tuning, we design a simple but effective two-step reformulation: removing self-talk behaviors and reordering sub-conclusions. A small auxiliary model is trained to perform this reformulation, incurring minimal computational overhead. Extensive experiments demonstrate that \ours{} consistently disrupts distillation across student models of different sizes and types on various reasoning benchmarks. For instance, when training on reformulated traces, even the performance of a large 32B student model decreases from 54.17 to 46.88 on AIME 2024, corresponding to a 13.5\% degradation.
\end{abstract}

\begin{figure*}[ht]
    \centering
    \includegraphics[width=0.8\textwidth]{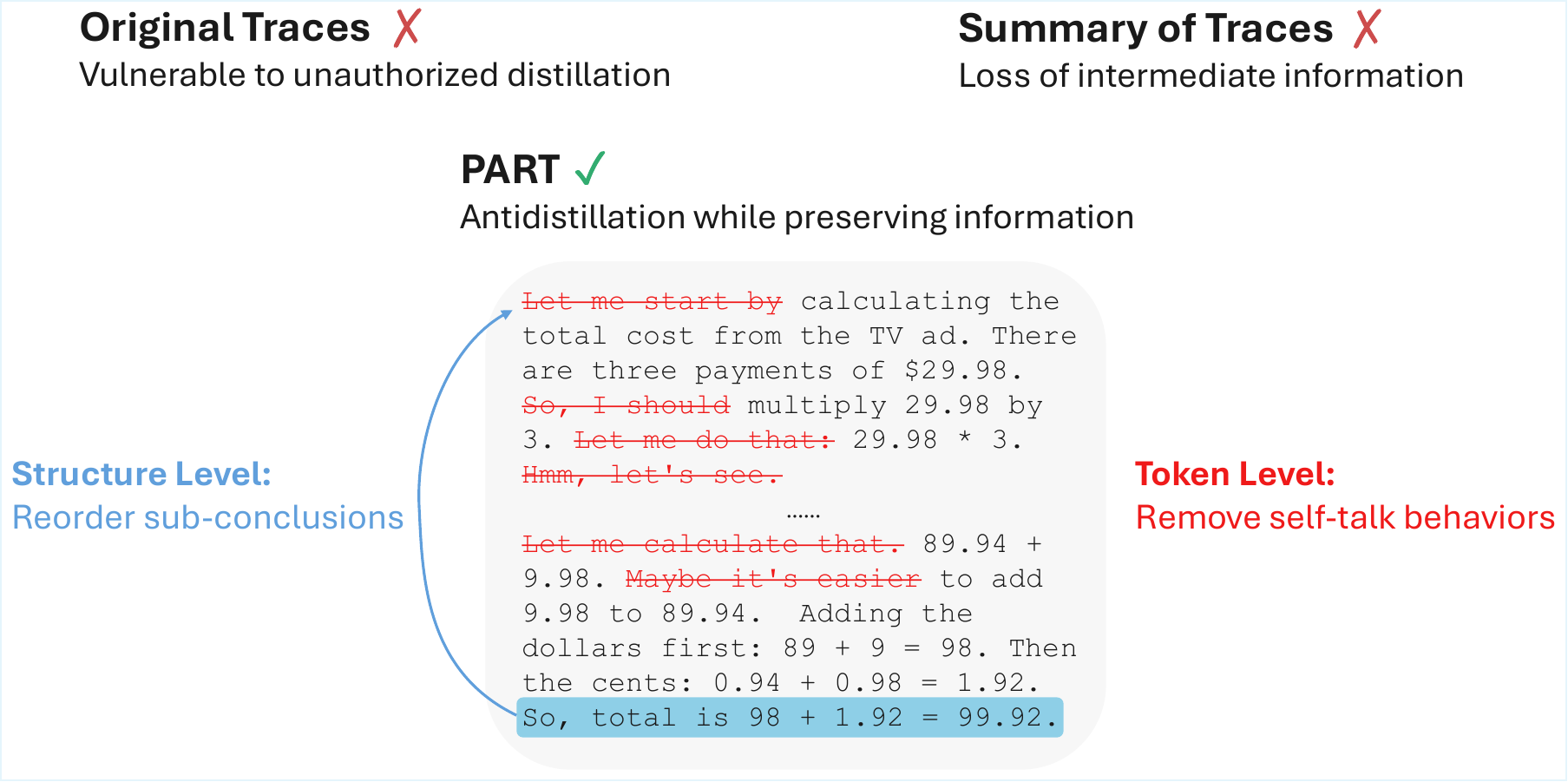}
    \caption{Overview of \ours{}. Directly exposing original reasoning traces leaves them vulnerable to unauthorized distillation, whereas providing only summaries deprives users of the information contained in the reasoning process. \ours{} introduces an information-preserving antidistillation approach through reformulation at both the token level and the structural level.}
\end{figure*}

\section{Introduction}

Large language models (LLMs) have recently achieved remarkable progress in domains such as mathematics and programming, largely driven by the use of long reasoning traces under test-time scaling \cite{o1, R1}. Beyond enhancing performance, these reasoning traces also allow users to gain insights into the model’s problem-solving process, thereby improving interpretability and trustworthiness of LLMs' response. However, exposing original reasoning traces makes them highly vulnerable to unauthorized distillation. It has been shown that supervised fine-tuning (SFT) on as few as ten of thousands of reasoning traces suffices for student models to attain comparable reasoning capabilities, leading to intellectual property leakage \cite{o1replicationjourney}. 

To mitigate this risk, existing proprietary model providers often adopt restrictive strategies to protect their reasoning traces. Common practices include either eliminating access to the reasoning trace or only revealing a condensed summary. While such strategies could prevent distillation, they hinder users from obtaining valuable information in reasoning traces. Recent works have explored antidistillation by controlling the sampling process or fine-tuning the teacher model \cite{antidistillationsampling,doge}. However, these approaches either compromise the performance of the teacher model or incur training costs for large teacher models. 

To address this issue, we introduce \ours{}, an information-\textbf{P}reserving \textbf{A}ntidistillation \textbf{R}eformulation of reasoning \textbf{T}races. The key insight of \ours{} is that the way models acquire reasoning ability through SFT differs from how humans comprehend reasoning processes. Reasoning traces that are interpretable for humans may not be suitable for distillation \cite{human}. Leveraging this discrepancy, we could defend distillation while preserving information. Concretely, we reformulate reasoning traces in two steps, modifying them at both the token level and the structural level, and we further train a small auxiliary model to perform this reformulation with minimal computational overhead.

At the token level, different tokens contribute unequally to parameter updates during SFT: tokens with lower predicted probabilities induce large gradients. Our analysis of student models’ learning dynamics on teacher-generated sequences reveals that these low-probability tokens contain many self-talk behaviors such as “Hmm,” “Wait,” and “Let’s”. While such expressions do not carry reasoning-related information, they drive substantial loss reduction during training, acting as the “useful tokens” discussed in prior studies \cite{rho}. Removing these expressions therefore disrupts distillation without sacrificing informational content.

At the structural level, prior work has shown that the structural perturbations to reasoning traces can significantly impact distillation \cite{structure}. To construct an information-preserving structural perturbation, we exploit the difference between reasoning understanding and generation. Humans do not require strict process-then-conclusion order to comprehend reasoning. Instead, it is common to use conclusion-before-process structures, such as presenting a lemma before its proof in mathematics. In contrast, limited by single-step computation, it's hard for LLMs to directly generate the correct conclusion without intermediate reasoning steps. Based on this difference, we reorder reasoning traces by placing sub-conclusions ahead of their corresponding reasoning steps. This reordering perturbs the structural patterns on which distillation relies, thereby weakening its effectiveness while maintaining human interpretability.

To verify that \ours{} effectively preserves information after reformulation, we evaluated the reformulated reasoning traces produced against the original reasoning traces from three perspectives: lexical similarity, semantic similarity, and human judgment. For lexical similarity, we segment the original reasoning traces into fragments and compute the match ratio in the reformulated reasoning traces using fuzzy matching. Experimental results show that across all similarity thresholds, \ours{} consistently outperforms the summary-based method. For semantic similarity, we employed Qwen3-Embedding-4B \cite{qwen3embedding} to map reasoning traces into embeddings and compute the match ratio by using the embeddings of the original traces as queries. Results demonstrate that 90.1\% of queries matched the reformulated reasoning traces generated by \ours{}, while only 7.3\% matched those produced by the summary-based method. Furthermore, in a user study on perceived informativeness, participants generally judged the information in \ours{} reformulations to be comparable to the originals, while clearly preferring \ours{} over the summary-based method for providing richer information.

To evaluate the antidistillation capability of \ours{}, we conducted experiments on student models of different sizes and types, comparing their performance when distilled with the original reasoning traces versus the reformulated reasoning traces. Results show that models trained on data reformulated by \ours{} suffer significant degradation across mathematics, coding, and scientific question answering benchmarks. \ours{} demonstrates stable effectiveness across varying model sizes and dataset scale. Notably, even a 32B student model exhibited a performance drop from 54.17 to 46.8 on AIME 2024, corresponding to a 13.5\% degradation.

The contributions of this paper are summarized as follows:

\begin{itemize}
    \item We propose \ours{}, a simple but effective reasoning trace reformulation method that disrupts distillation while preserving information. We validate that our approach successfully retains the information contained in the original reasoning traces from multiple perspectives, including lexical similarity, semantic similarity, and human judgment.
    \item We conducted extensive distillation experiments and demonstrate that our method effectively degrades the performance of distilled models across student models up to 32B parameters, varying amounts of training data, and diverse downstream tasks.
    \item We will release the code and data to facilitate future research on antidistillation and reasoning trace reformulation.
\end{itemize}

\section{Method}
\subsection{Problem Formulation}

Knowledge distillation aims to leverage a strong teacher model $T$ to guide a lightweight student model $S$, transferring the teacher's capabilities to the student \cite{kdsurvey, kdsurveyllm}. A common approach to distilling large language models is supervised fine-tuning (SFT) on the data generated by the teacher. The student model is optimized to maximize the log-likelihood of the teacher's output $y$ conditioned on query $q$:

\begin{equation}
\mathcal{L}_{\mathrm{SFT}}(\theta_S) = - \frac{1}{T} \sum_{t=1}^T \log p_{\theta_S}(y_t \mid y_{<t}, q)
\end{equation}

For reasoning models, each output sequence $y = (r, a)$ consists of a reasoning trace $r$ and a final answer $a$. To interfere with distillation, we consider a transformation $\mathcal{T}: r \mapsto r'$ that rewrites the reasoning trace. We keep the final answer unchanged, because it conveys the task outcome from which users extract the final result. Existing proprietary models often adopt restrictive disclosure strategies to prevent unauthorized distillation by others. For example, they omit the reasoning trace entirely or present only a high-level summary. However, these ways cause substantial information loss for users. 

Our goal is therefore to design a transformation $\mathcal{T}$ that meets the following two objectives: 
\begin{itemize}
    \item Interfere with distillation. Make the distilled model $S_{\mathcal{D}_\mathcal{T}}$, which is trained on the modified dataset $\mathcal{D}_\mathcal{T} = \{(q_i, \mathcal{T}(r_i), a_i)\}_{i=1}^N$, yield degraded downstream performance $\mathrm{Perf}(S_{\mathcal{D}_\mathcal{T}})$.
    \item Information preservation. Ensure that the modified reasoning trace $ \mathcal{T}(r_i)$ remains human-readable and preserves as much useful information as possible in each $r_i$, so that it stays interpretable and useful for human readers.
\end{itemize}

Formally, this trade-off can be posed as the constrained optimization problem:
\begin{equation}
\begin{aligned}
\arg\min_{\mathcal{T}} \quad & \mathrm{Perf}(S_{\mathcal{D}_\mathcal{T}}) \\
\text{s.t.} \quad & \mathrm{Sim}(r_i, \mathcal{T}(r_i)) > \tau, \quad \forall i
\end{aligned}
\end{equation}
where $\mathrm{Sim}(r_i, \mathcal{T}(r_i))$ is a similarity measure between the original and rewritten reasoning trace, and $\tau$ is a similarity threshold ensuring that $r'$ remains sufficiently faithful to $r$ from a human reader's perspective.

\subsection{Reasoning Trace Reformulation}

To construct an information-preserving antidistillation reformulation method, the key lies in identifying the differences between how LLMs learn reasoning through SFT and how humans comprehend reasoning traces. Prior studies have shown that reasoning traces that are easily understood by humans are not necessarily suitable for distillation; in fact, manually annotated chain-of-thought data sometimes perform poorly in distillation \cite{human}. Leveraging this discrepancy, we design a reformulation method from two complementary perspectives: the token level and the structural level.

\textbf{1. Removing self-talk behaviors}

\begin{figure}[t]
    \centering
    \subfigure[]{\label{subfig:prob_a}
        \includegraphics[width=0.47\textwidth]{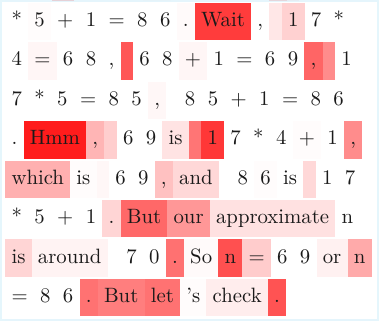}
    }
    \subfigure[]{\label{subfig:prob_b}
        \includegraphics[width=0.41\textwidth]{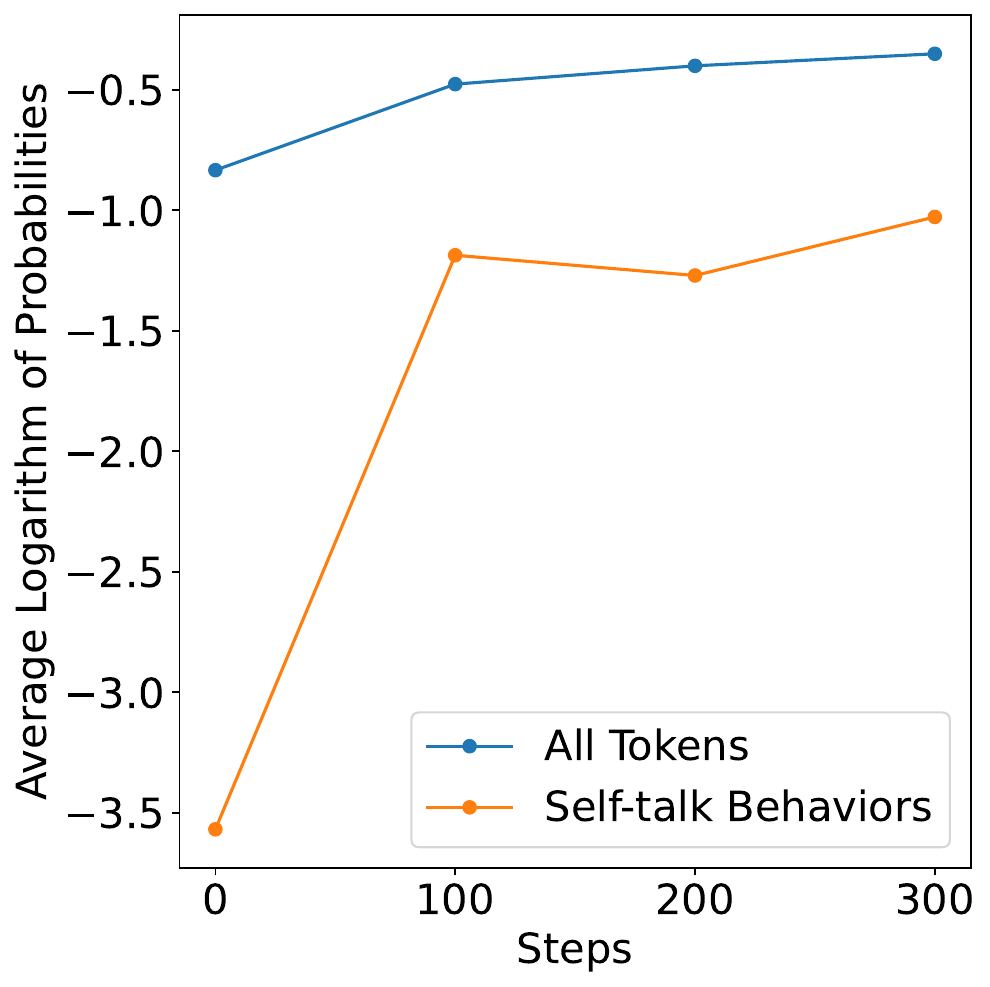}
    }
    \caption{Predicted probabilities of the student model on teacher-generated reasoning traces. (a) Visualization of token-level predicted probabilities, where deeper red indicates lower probabilities. Teacher-generated traces exhibit frequent self-talk behaviors, which conveys little reasoning content yet receives low probabilities. (b) Tracking the probabilities of self-talk-behavior tokens across training stages reveals that they remain persistently lower than the average probabilities, suggesting that these semantically uninformative expressions exert disproportionate influence on gradient updates.}
    \label{fig:prob}
\end{figure}

At the token level, we first analyze how different tokens contribute to the SFT from the gradient perspective.
For a single time step $t$, $y_t$ denotes the ground-truth token at position $t$, and define the logits vector is denoted as $z^{(t)} \in \mathbb{R}^V$ over the vocabulary of size $V$. The token-level loss is
\begin{equation}
\mathcal{L}^{(t)}(\theta) = -\log p^{(t)}_{y_t} = -\log(\frac{e^{z^{(t)}_i}}{\sum_{j=1}^V e^{z^{(t)}_j}}).
\end{equation}

The gradient of this loss with respect to the logits is
\begin{equation}
\nabla_{z^{(t)}} \mathcal{L}^{(t)} = p^{(t)} - e_{y_t},
\end{equation}
where $e_{y_t}$ is the one-hot indicator vector for the target token. 

The squared $\ell_2$-norm of the gradient vector is
\begin{equation}
\big\| \nabla_{z^{(t)}} \mathcal{L}^{(t)} \big\|_2^2 
= \sum_{i=1}^V \left(p^{(t)}_i - e_{y_t,i}\right)^2 
= \sum_{i=1}^V (p^{(t)}_i)^2 + 1 - 2p^{(t)}_{y_t}
\end{equation}

This expression reveals a direct dependence on the predicted probability of the correct token $p^{(t)}_{y_t}$. When $p^{(t)}_{y_t} \to 1$, the gradient approaches zero. When $p^{(t)}_{y_t}$ is small, the gradient norm grows and signals a strong update. This indicates that tokens that the model already predicts with high confidence contribute negligible gradients and quickly fade from influencing optimization. In contrast, low-probability (i.e., poorly predicted) tokens dominate the effective training signal, guiding the model to adjust its parameters toward correcting these mistakes.

Based on the analysis, we examine the predicted probabilities of the student model on teacher-generated reasoning traces. \cref{subfig:prob_a} visualizes the predicted probabilities on a segment of a teacher-generated trace, where deeper red indicates lower probabilities. We observe that low-probability tokens contain many self-talk behaviors, a phenomenon where the model often speaks in the first person and employs colloquial expressions such as “hmm” and “wait”. These expressions contain little information relevant to reasoning. However, the student model assigns low probabilities to such tokens, which results in large gradient.

We further track the predicted probabilities of representative tokens like “Hmm” and “Wait” across different training stages. As shown in \cref{subfig:prob_b}, these tokens exhibit persistently lower probabilities than the average token, which implies that such semantically uninformative expressions exert disproportionate influence during parameter updates. Previous studies have also explored the different influences of tokens during training. \cite{rho} identified distinct loss patterns across tokens in pretraining: some tokens consistently maintain high or low loss, while only a subset exhibits significant loss reduction and are regarded as useful tokens. Tokens associated with self-talk behaviors demonstrate a similar pattern, indicating their impact on training.

To leverage this, we rewrite the reasoning traces to remove self-talk behaviors. This modification incurs negligible information loss, while deliberately perturbing the gradients associated with low-probability tokens, thereby affecting the distillation process. 

\textbf{2. Reordering the sub-conclusions}

At the sequence level, LLMs learn to imitate the overall logical structure of reasoning traces in order to perform reasoning. \cite{structure} demonstrates that structural perturbations to reasoning traces have a substantial impact on the performance of distilled models. However, their methods focused on operations such as randomly shuffling or deleting steps, or inserting irrelevant steps, which severely compromise human readability. 

To design a form of structural perturbation that preserves readability for humans, we exploit the difference between generating a reasoning process and understanding a reasoning process. Reasoning generation proceeds in a strictly sequential manner: intermediate steps must be generated before reaching the conclusion. In contrast, comprehension does not require this order; humans often prefer presenting the conclusion first, followed by the supporting process. For example, in mathematical reasoning, lemmas are often stated prior to their proofs, and in academic writing, abstracts precede detailed methods and results. This conclusion-before-process structure can even enhance human understanding of reasoning.

For LLMs, reordering sub-conclusions breaks the chain-of-thought structure of original reasoning traces. Since the computational capacity per step is bounded, LLMs without chain-of-thought can only solve problems of limited complexity \cite{cot}. The models struggle to directly generate correct conclusions without generating reasoning processes first. This limitation makes it difficult for an LLM to distill reasoning traces with conclusion-before-process order.

Leveraging this asymmetry, we rewrite reasoning traces by reordering them into a conclusion-before-process structure. Specifically, we prompt GPT-4o \cite{gpt4o} to reformulate reasoning traces in a chain-of-thought style, where sub-conclusions are first summarized and then placed before their associated reasoning process.

\subsection{Training a Compact Reformulation Model}

In practical applications, it is desirable to minimize the cost of reformulation so as to mitigate its impact on the inference service. To this end, we train a compact model for reformulation. Each original reasoning trace is divided into multiple segments, which are reformulated individually and then concatenated back into a complete trace. We fine-tune Qwen2.5-1.5B-Instruct \cite{qwen25} using paired data consisting of original reasoning traces and their rewritten counterparts generated by GPT-4o. Compared with advanced reasoning models such as DeepSeek-R1, the reformulation model introduces less than 1\% additional parameters and incurs only about 4\% extra computational overhead.

\cref{sec:example} is an example of reformulation generated by our compact reformulation model. Despite its relatively small size, the model effectively accomplishes the reformulation task, successfully performing removal and reordering while preserving the information. In \cref{sec:quantity}, we present a quantitative evaluation of the reformulation model, demonstrating its ability to achieve both antidistillation with information preservation.

\section{Quality of Reformulated Reasoning Traces}

When reformulating reasoning traces to defend distillation, it is necessary not only to ensure the impact on the performance of distilled models but also to maintain the quality of the reformulated traces. In extreme cases, completely nonsensical reasoning traces would indeed prevent successful distillation, but they would also be unreadable to users and thus fail to convey any useful information. To this end, we evaluate the quality of reformulated reasoning traces using three complementary approaches: lexical similarity, semantic similarity, and human judgment. We compare the quality of traces reformulated by \ours{} with that of the original traces and segment-level summaries. These methods verify that our reformulations can disrupt distillation while still preserving the information of the original reasoning traces, thereby ensuring usability for users.  

\begin{figure}[t]
    \centering
    \subfigure[]{\label{subfig:lexical}
        \includegraphics[width=0.41\textwidth]{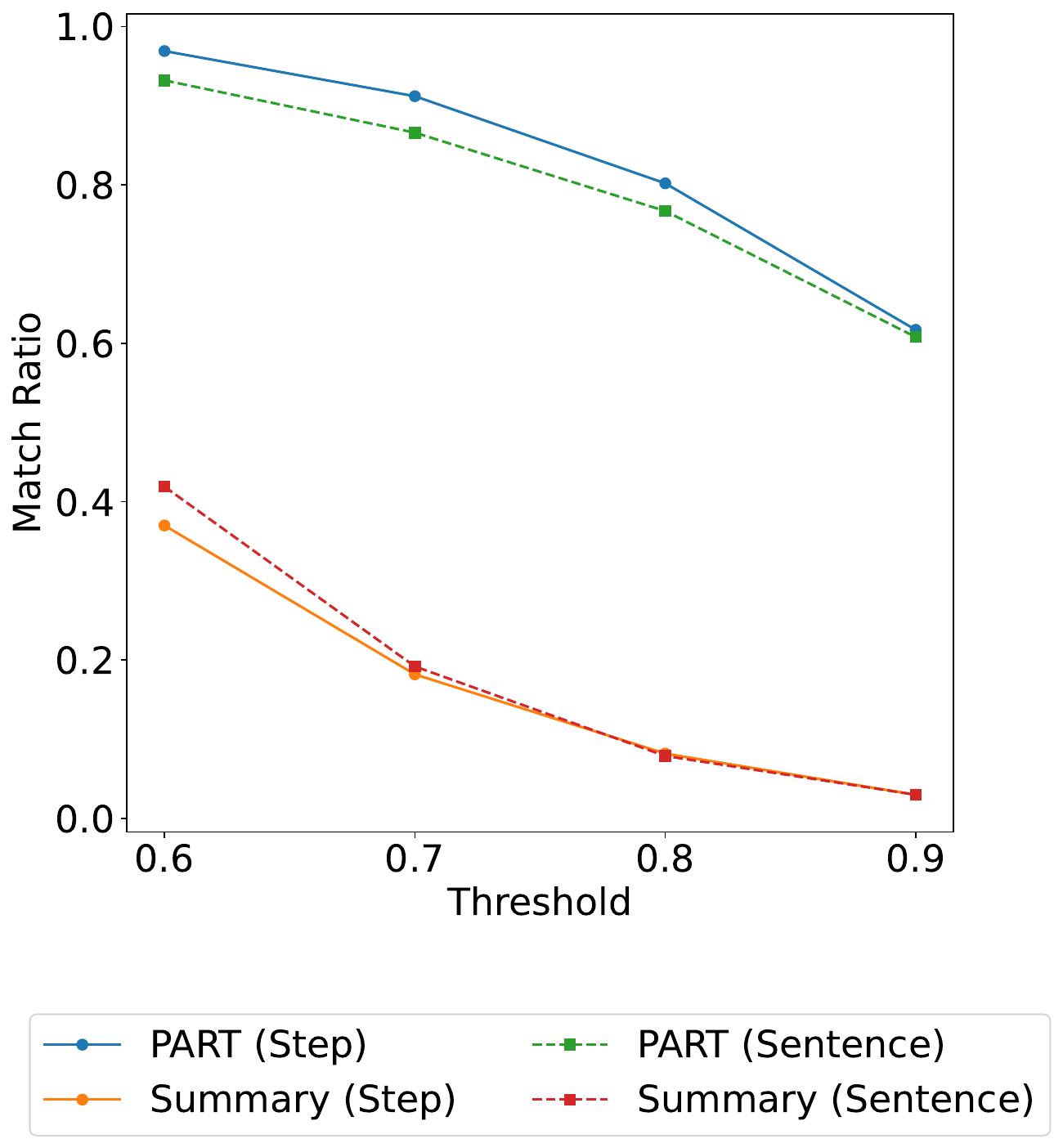}
    }
    \subfigure[]{\label{subfig:preference}
        \includegraphics[width=0.45\textwidth]{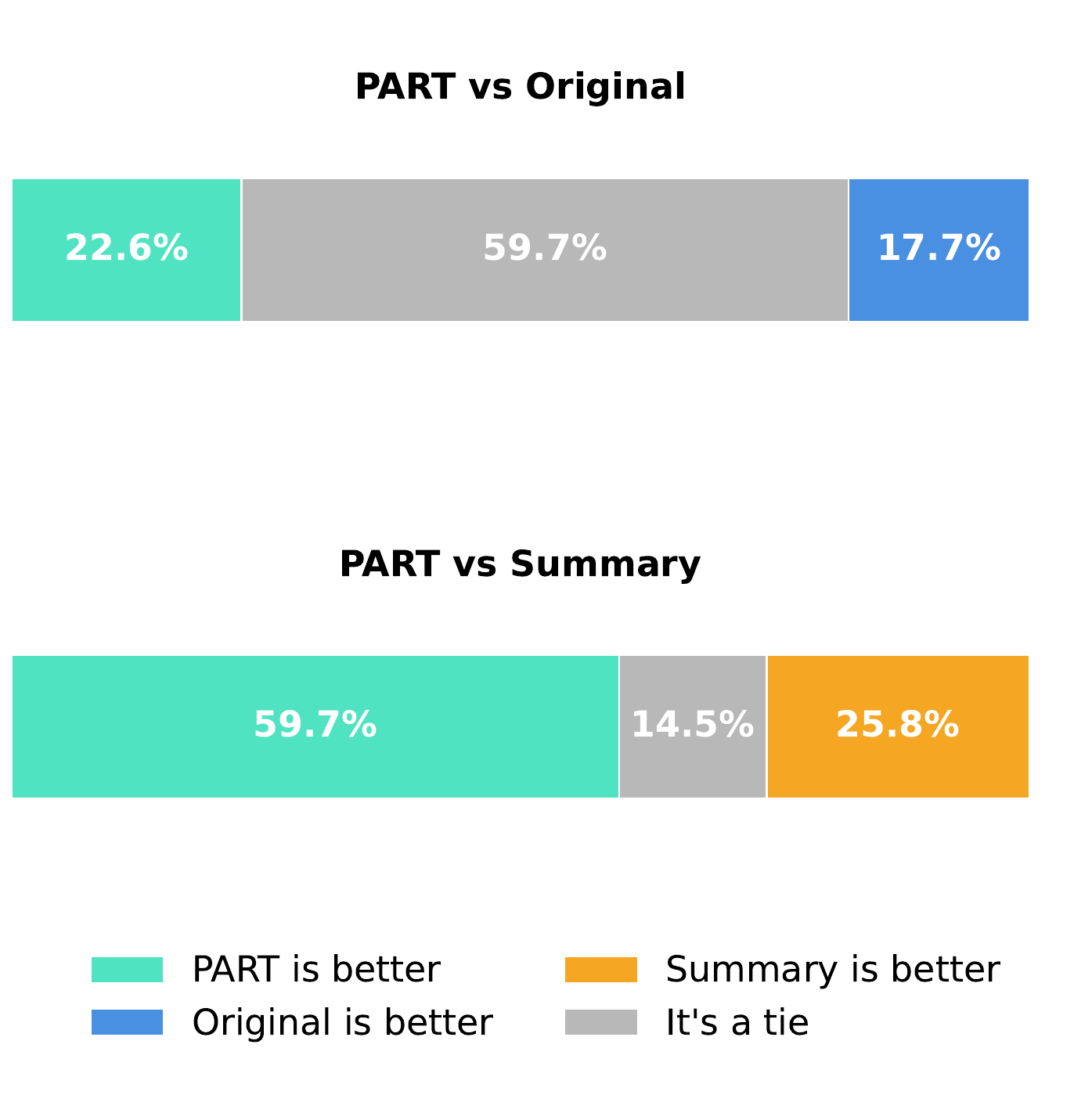}
    }
    \caption{ (a) Match ratios under different lexical similarity score thresholds. \ours{} achieves significant higher match ratios than the summary-based method at both step and sentence levels. (b) Human judgment about informativeness. Compared to original reasoning traces, \ours{} is judged similarly informative; compared to the summary-based reformulation, \ours{} is clearly preferred in terms of the informativeness.}
    \label{fig:similarity}
\end{figure}

\subsection{Lexical Similarity}  

A straightforward way to compare the lexical similarity between the original and reformulated reasoning traces is to perform fuzzy matching at the segment level. Specifically, we split the original reasoning trace into sentences or steps and check whether they can be successfully matched within the reformulated reasoning trace. We employ the \texttt{partial\_ratio\_alignment} function from the \texttt{RapidFuzz} library to calculate the similarity score, which first performs substring matching and then computes the normalized Indel similarity based on edit distance.

As shown in \cref{subfig:lexical}, we compute the match ratio under different similarity score thresholds. Whether the matching is conducted at the step or sentence level, \ours{} exhibits a remarkably high match ratio, substantially surpassing summary-based approaches across different thresholds. Examples of matched text pairs under different thresholds are shown in \cref{sec:thres}. At a threshold of 0.7, \ours{} achieved a match ratio of 91\%, whereas the summary-based method achieved only 18\%.  This indicates that \ours{} is able to interfere with distillation through only minimal textual modifications.

\subsection{Semantic Similarity}

To compare the semantic similarity of reasoning traces, we employed Qwen3-Embedding-4B \cite{qwen3embedding} to map reasoning traces into text embedding, under the assumption that semantically closer reasoning traces should yield higher embedding similarity. We treated the original reasoning traces as queries, and the reformulated reasoning traces produced by \ours{} and the summary-based approach as candidate documents.

Experimental results show that 97.4\% of queries successfully matched their corresponding reformulated reasoning traces. Specifically, 90.1\% of queries matched the reasoning traces reformulated by \ours{}, whereas only 7.3\% matched those reformulated by the summary-based method. Moreover, the average cosine similarity between the original reasoning traces and those reformulated by \ours{} reached 0.950, compared to 0.889 for the summary-based method. These results demonstrate that \ours{} achieves superior semantic similarity to the original reasoning traces.

\subsection{Human Judgment}

To assess user perceptions of reformulated reasoning traces, we conducted a questionnaire study on their perceived informativeness. We sampled 50 original reasoning traces, each paired with two reformulated versions: one generated by \ours{} and one by a summary-based method. Each participant evaluated four pairs of traces, indicating their preference (“A is better,” “B is better,” or “Tie”). Two pairs compared original traces with those reformulated by \ours{}, and the other two compared summary-based reformulations with those from \ours{}.

We collected 31 completed questionnaires, from which we obtained 124 comparisons. As shown in \cref{subfig:preference}, when comparing reasoning traces produced by \ours{} with the original traces, most participants judged the information content to be comparable. In contrast, when comparing \ours{} with the summary-based method, participants showed a clear preference for \ours{} in terms of the richness of information provided.

\section{Experiments}

\subsection{Setup}
\label{sec:setup}

To assess the impact of our reasoning trace reformulation method on the effectiveness of distillation, we distill student models using both the original reasoning traces and the rewritten reasoning traces, and compare their performance. 

\paragraph{Training Setup.} We experiment with student models of different sizes and families. Specifically, we follow DeepSeek-R1 \cite{R1} in selecting the base models: Qwen2.5-Math-1.5B, Qwen2.5-Math-7B \cite{qwenmath}, Qwen2.5-14B, Qwen2.5-32B \cite{qwen25}. In addition, we also examine distillation with an instruct model as the student model, for which we use Qwen2.5-7B-Instruct. Since the Qwen2.5-Math models only support a maximum context length of 4K tokens, we extend their context window by setting the \texttt{rope\_theta} parameter to 1,000,000 following \cite{acereason}. For distillation data, we used the Bespoke-Stratos-17k \cite{bespoke_stratos} and OpenThoughts-114k datasets \cite{openthoughts}. We adopt the Llama-Factory \cite{llamafactory} framework to perform SFT.

\paragraph{Evaluation Setup.} For evaluation, we evaluate the distilled models on MATH-500 \cite{math,math500}, AIME 2024 \cite{AIME2024}, LiveCodeBench v2 \cite{livecodebench}, and GPQA-Diamond \cite{gpqa}, covering tasks in mathematical reasoning, code generation, and scientific question answering. To obtain more reliable estimates of pass@1 accuracy, we sample multiple responses per query, thereby reducing variance in the results. Details of training and evaluation setup are provided in \cref{sec:hyperparameter}.

\subsection{Results}

\begin{table}[t]
\caption{Performance of distilled models on various benchmarks. “MATH500” refers to MATH-500, “AIME24” to AIME 2024, “LCBv2” to LiveCodeBench v2, and “GPQA-D” to GPQA-Diamond. More negative values of $\Delta$ indicate stronger antidistillation effects.}
\label{tab:results}
\vspace{3pt}
\begin{center}
\begin{tabular}{llccccc}
\toprule
Student Model & Data & MATH500 & AIME24 & LCBv2 & GPQA-D & Average \\ \midrule
\addlinespace[0.5ex]
\multicolumn{7}{c}{\textbf{Training Data: Bespoke-Stratos-17k}} \\
\addlinespace[0.5ex]
\multirow{2}{*}{Qwen2.5-Math-1.5B} & original & 72.55 & 15.00 & 12.23& 29.80 & 32.40 \\
 & \ours{} & 59.05 & 8.75 & 9.88 & 25.88 & 25.89 \\
\rowcolor{blue!20} \cellcolor{white}
 & $\Delta$ & -13.50 & -6.25 & -2.35 & -3.92 & -6.51 \\
Qwen2.5-Math-7B & original & 88.95 & 32.71 & 34.88 & 43.18 & 49.93 \\
 & \ours{} & 80.00 & 22.08 & 28.96 & 38.00 & 42.26 \\
\rowcolor{blue!20} \cellcolor{white}
 & $\Delta$ & -8.95 & -10.63 & -5.92 & -5.18 & -7.67 \\
Qwen2.5-14B & original & 90.60 & 43.75 & 55.58 & 53.28 & 60.80 \\
 & \ours{} & 82.25 & 25.83 & 43.98 & 46.97 & 49.76 \\
\rowcolor{blue!20} \cellcolor{white}
 & $\Delta$ & -8.35 & -17.92 & -11.60 & -6.31 & -11.05 \\
Qwen2.5-32B & original & 92.65 & 54.17 & 70.99 & 61.24 & 69.76 \\
 & \ours{} & 89.65 & 46.88 & 62.38 & 55.68 & 63.65 \\
\rowcolor{blue!20} \cellcolor{white}
 & $\Delta$ & -3.00 & -7.29 & -8.61 & -5.56 & -6.12 \\
Qwen2.5-7B-Instruct & original & 83.05  & 21.04 & 36.64 & 43.31 & 46.01 \\
 & \ours{} & 70.85 & 12.29 & 27.84 & 32.32 & 35.83 \\
\rowcolor{blue!20} \cellcolor{white}
 & $\Delta$ & -12.20 & -8.75 & -8.80 & -10.99 & -10.19 \\
 \addlinespace[0.5ex]
 \multicolumn{7}{c}{\textbf{Training Data: OpenThoughts-114k}} \\
 \addlinespace[0.5ex]
 Qwen2.5-Math-7B & original & 90.40 & 46.67 & 41.98 & 45.20 & 56.06 \\
 & \ours{} & 78.50 & 28.54 & 30.87 & 36.49 & 43.60 \\
 \rowcolor{blue!20} \cellcolor{white}
 & $\Delta$ & -11.90 & -18.13 & -11.11 & -8.71 & -12.46 \\
 \bottomrule
\end{tabular}
\end{center}
\end{table}

\cref{tab:results} reports the performance of distilled models across different benchmarks. It shows that, regardless of student model size or benchmark, \ours{} consistently leads to a significant degradation in the performance of distilled models, thereby providing an effective defense against distillation. For example, even the performance of a large 32B student model decreases from 54.17 to 46.88 on AIME 2024, corresponding to a 13.5\% degradation.

For the student model in reasoning distillation, some studies adopt a base model \cite{R1, acereason}, while others use an instruct model \cite{bespoke_stratos}. We conducted experiments on both choices of student models. When using Qwen2.5-Math-7B as the student model, the average score decreases by 7.67, and when using Qwen2.5-7B-Instruct as the student model, the score decreases by 10.19. This demonstrates that \ours{} is effective against different types of student models.

\subsection{Effectiveness of the Reformulation Model}
\label{sec:quantity}

To evaluate the generalization capability of the reformulation model, we trained a 1.5B reformulation model on reformulated data generated by GPT-4o using the Bespoke-Stratos-17k dataset and applied it to reformulate OpenThoughts-114k. As shown in \cref{tab:results}, the reasoning traces produced by this small reformulation model also effectively defend against distillation: student models distilled on the reformulated data exhibit significant performance degradation.

We further evaluated the quality of traces generated by the reformulation model. For lexical similarity, the match ratio reached 88\% under a threshold of 0.7. For semantic similarity, the average cosine similarity was 0.94. These similarity metrics are close to those obtained with GPT-4o reformulations and substantially higher than those of the summary-based method. This demonstrates that our reformulation model is also effective in preserving information during reformulation.

\subsection{Detectability}

An additional property of \ours{} is detectability. Similar to LLM watermarking, models distilled on \ours{}-reformulated data can be distinguished from those trained on original reasoning traces. Due to the removal of self-talk behaviors, the distributional patterns of the data undergo significant changes. We computed the term frequency of keywords related to self-talk behaviors: in the original traces, the average frequency of such keywords reached 2.9\%, whereas in the reformulated traces it dropped to only 0.4\%. Leveraging this substantial discrepancy, even a simple classifier based on a term-frequency threshold is sufficient to achieve separation, yielding an F1 score of 0.93 and a true-positive rate of 88.3\% at a 1\% false-positive rate (TPR@FPR).

\begin{figure}[t]
    \centering
    \subfigure[]{\label{subfig:model_size}
        \includegraphics[width=0.45\textwidth]{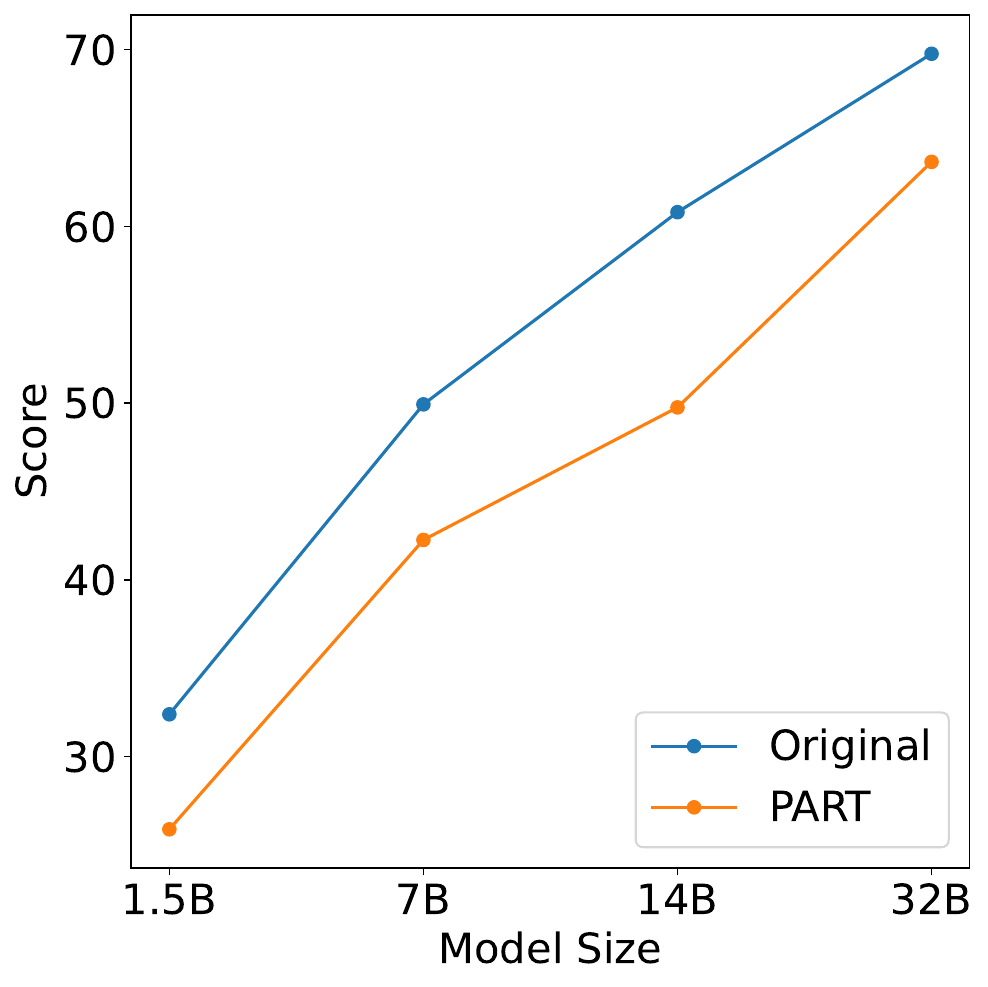}
    }
    \subfigure[]{\label{subfig:data_scale}
        \includegraphics[width=0.45\textwidth]{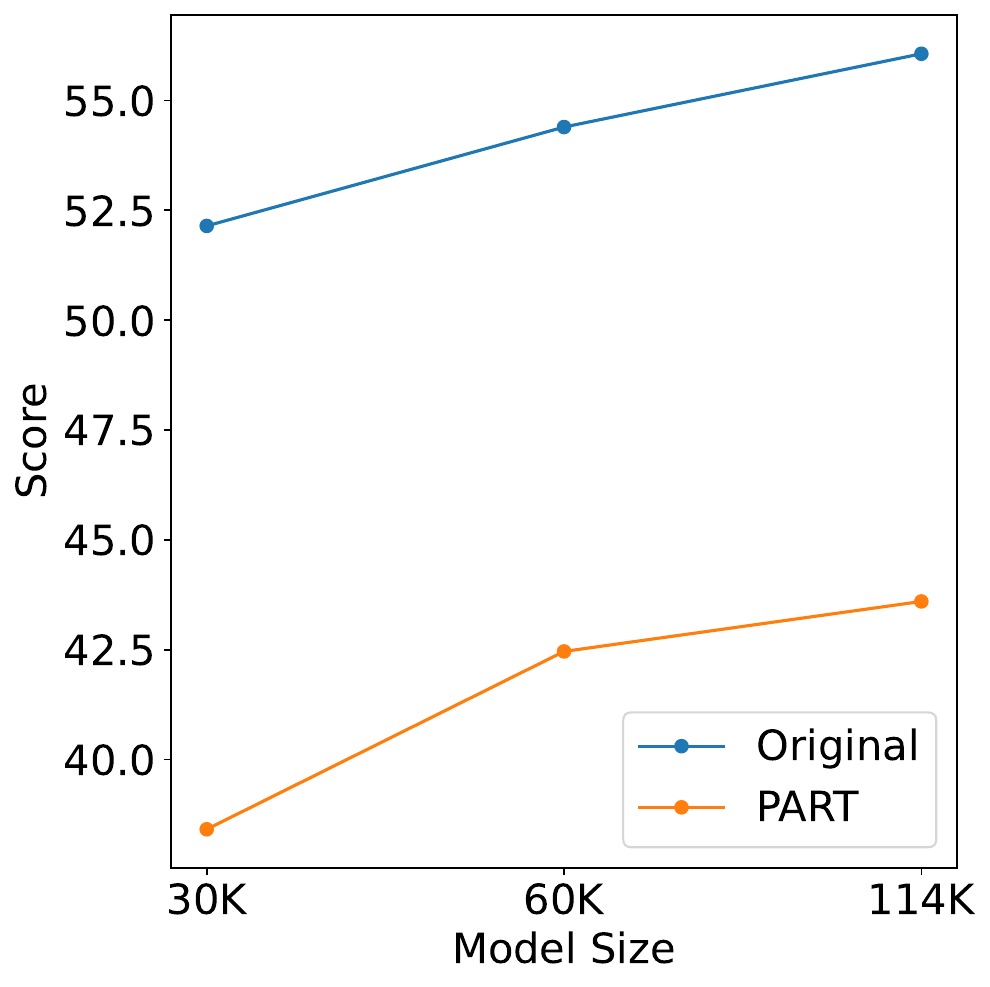}
    }
    \caption{Performance comparison (a) across different student model sizes and (b) across different data scales of distilled models trained on original versus reformulated traces. Across both factors, \ours{} leads to consistent performance degradation of the distilled models, demonstrating its effectiveness as an antidistillation approach.}
    \label{fig:scale}
\end{figure}

\subsection{Robustness to Data Scale}

To evaluate the effectiveness of \ours{} under varying amounts of training data, we sampled subsets of different sizes from the OpenThoughts-114K dataset and its corresponding reformulated traces. As shown in \cref{subfig:data_scale}, \ours{} consistently led to a significant degradation in distilled model performance across different data scales. Notably, models trained on a large number of reformulated traces still underperformed compared to those trained on only a smaller number of original traces. This finding indicates that it is costly to collect more data to offset the impact of \ours{}.

\section{Conclusion}

We presented \ours{}, an information-preserving reformulation of reasoning traces for antidistillation. Leveraging the difference between how humans comprehend reasoning process and how LLMs acquire reasoning ability via supervised fine-tuning, \ours{} applies two simple but effective steps: removing self-talk tokens and reorder sub-conclusions before their supporting process. Across lexical, semantic, and human-judgment evaluations, \ours{} retains the information of original traces, substantially outperforming summary-based method. Distillation experiments show consistent degradation for student models trained on reformulated traces across various benchmarks, robust to model size and data scale. Overall, \ours{} offers a practical method to balance interpretability with protection of model intellectual property.

\bibliography{neurips_2025}

\newcommand{\etalchar}[1]{$^{#1}$}
\begin{thebibliography}{DAGY{\etalchar{+}}25}

\bibitem[CSW{\etalchar{+}}25]{human}
Xinghao Chen, Zhijing Sun, Guo Wenjin, Miaoran Zhang, Yanjun Chen, Yirong Sun, Hui Su, Yijie Pan, Dietrich Klakow, Wenjie Li, and Xiaoyu Shen.
\newblock Unveiling the key factors for distilling chain-of-thought reasoning.
\newblock In Wanxiang Che, Joyce Nabende, Ekaterina Shutova, and Mohammad~Taher Pilehvar, editors, {\em Findings of the Association for Computational Linguistics: ACL 2025}, pages 15094--15119, Vienna, Austria, July 2025. Association for Computational Linguistics.

\bibitem[DAGY{\etalchar{+}}25]{R1}
DeepSeek-AI, Daya Guo, Dejian Yang, Haowei Zhang, Junxiao Song, Ruoyu Zhang, Runxin Xu, Qihao Zhu, Shirong Ma, Peiyi Wang, Xiao Bi, Xiaokang Zhang, Xingkai Yu, Yu~Wu, Z.~F. Wu, Zhibin Gou, Zhihong Shao, Zhuoshu Li, Ziyi Gao, Aixin Liu, Bing Xue, Bingxuan Wang, Bochao Wu, Bei Feng, Chengda Lu, Chenggang Zhao, Chengqi Deng, Chenyu Zhang, Chong Ruan, Damai Dai, Deli Chen, Dongjie Ji, Erhang Li, Fangyun Lin, Fucong Dai, Fuli Luo, Guangbo Hao, Guanting Chen, Guowei Li, H.~Zhang, Han Bao, Hanwei Xu, Haocheng Wang, Honghui Ding, Huajian Xin, Huazuo Gao, Hui Qu, Hui Li, Jianzhong Guo, Jiashi Li, Jiawei Wang, Jingchang Chen, Jingyang Yuan, Junjie Qiu, Junlong Li, J.~L. Cai, Jiaqi Ni, Jian Liang, Jin Chen, Kai Dong, Kai Hu, Kaige Gao, Kang Guan, Kexin Huang, Kuai Yu, Lean Wang, Lecong Zhang, Liang Zhao, Litong Wang, Liyue Zhang, Lei Xu, Leyi Xia, Mingchuan Zhang, Minghua Zhang, Minghui Tang, Meng Li, Miaojun Wang, Mingming Li, Ning Tian, Panpan Huang, Peng Zhang, Qiancheng Wang, Qinyu Chen, Qiushi Du, Ruiqi Ge, Ruisong
  Zhang, Ruizhe Pan, Runji Wang, R.~J. Chen, R.~L. Jin, Ruyi Chen, Shanghao Lu, Shangyan Zhou, Shanhuang Chen, Shengfeng Ye, Shiyu Wang, Shuiping Yu, Shunfeng Zhou, Shuting Pan, S.~S. Li, Shuang Zhou, Shaoqing Wu, Shengfeng Ye, Tao Yun, Tian Pei, Tianyu Sun, T.~Wang, Wangding Zeng, Wanjia Zhao, Wen Liu, Wenfeng Liang, Wenjun Gao, Wenqin Yu, Wentao Zhang, W.~L. Xiao, Wei An, Xiaodong Liu, Xiaohan Wang, Xiaokang Chen, Xiaotao Nie, Xin Cheng, Xin Liu, Xin Xie, Xingchao Liu, Xinyu Yang, Xinyuan Li, Xuecheng Su, Xuheng Lin, X.~Q. Li, Xiangyue Jin, Xiaojin Shen, Xiaosha Chen, Xiaowen Sun, Xiaoxiang Wang, Xinnan Song, Xinyi Zhou, Xianzu Wang, Xinxia Shan, Y.~K. Li, Y.~Q. Wang, Y.~X. Wei, Yang Zhang, Yanhong Xu, Yao Li, Yao Zhao, Yaofeng Sun, Yaohui Wang, Yi~Yu, Yichao Zhang, Yifan Shi, Yiliang Xiong, Ying He, Yishi Piao, Yisong Wang, Yixuan Tan, Yiyang Ma, Yiyuan Liu, Yongqiang Guo, Yuan Ou, Yuduan Wang, Yue Gong, Yuheng Zou, Yujia He, Yunfan Xiong, Yuxiang Luo, Yuxiang You, Yuxuan Liu, Yuyang Zhou, Y.~X. Zhu,
  Yanhong Xu, Yanping Huang, Yaohui Li, Yi~Zheng, Yuchen Zhu, Yunxian Ma, Ying Tang, Yukun Zha, Yuting Yan, Z.~Z. Ren, Zehui Ren, Zhangli Sha, Zhe Fu, Zhean Xu, Zhenda Xie, Zhengyan Zhang, Zhewen Hao, Zhicheng Ma, Zhigang Yan, Zhiyu Wu, Zihui Gu, Zijia Zhu, Zijun Liu, Zilin Li, Ziwei Xie, Ziyang Song, Zizheng Pan, Zhen Huang, Zhipeng Xu, Zhongyu Zhang, and Zhen Zhang.
\newblock Deepseek-r1: Incentivizing reasoning capability in llms via reinforcement learning, 2025.

\bibitem[DSG{\etalchar{+}}24]{scalablewatermark}
Sumanth Dathathri, Abigail See, Sumedh Ghaisas, Po-Sen Huang, Rob McAdam, Johannes Welbl, Vandana Bachani, Alex Kaskasoli, Robert Stanforth, Tatiana Matejovicova, et~al.
\newblock Scalable watermarking for identifying large language model outputs.
\newblock {\em Nature}, 634(8035):818--823, 2024.

\bibitem[GMK{\etalchar{+}}25]{openthoughts}
Etash Guha, Ryan Marten, Sedrick Keh, Negin Raoof, Georgios Smyrnis, Hritik Bansal, Marianna Nezhurina, Jean Mercat, Trung Vu, Zayne Sprague, Ashima Suvarna, Benjamin Feuer, Liangyu Chen, Zaid Khan, Eric Frankel, Sachin Grover, Caroline Choi, Niklas Muennighoff, Shiye Su, Wanjia Zhao, John Yang, Shreyas Pimpalgaonkar, Kartik Sharma, Charlie Cheng-Jie Ji, Yichuan Deng, Sarah Pratt, Vivek Ramanujan, Jon Saad-Falcon, Jeffrey Li, Achal Dave, Alon Albalak, Kushal Arora, Blake Wulfe, Chinmay Hegde, Greg Durrett, Sewoong Oh, Mohit Bansal, Saadia Gabriel, Aditya Grover, Kai-Wei Chang, Vaishaal Shankar, Aaron Gokaslan, Mike~A. Merrill, Tatsunori Hashimoto, Yejin Choi, Jenia Jitsev, Reinhard Heckel, Maheswaran Sathiamoorthy, Alexandros~G. Dimakis, and Ludwig Schmidt.
\newblock Openthoughts: Data recipes for reasoning models, 2025.

\bibitem[GYMT21]{kdsurvey}
Jianping Gou, Baosheng Yu, Stephen~J Maybank, and Dacheng Tao.
\newblock Knowledge distillation: A survey.
\newblock {\em International journal of computer vision}, 129(6):1789--1819, 2021.

\bibitem[HBK{\etalchar{+}}21]{math}
Dan Hendrycks, Collin Burns, Saurav Kadavath, Akul Arora, Steven Basart, Eric Tang, Dawn Song, and Jacob Steinhardt.
\newblock Measuring mathematical problem solving with the math dataset.
\newblock In {\em Thirty-fifth Conference on Neural Information Processing Systems Datasets and Benchmarks Track (Round 2)}, 2021.

\bibitem[HZL{\etalchar{+}}24]{o1replicationjourney}
Zhen Huang, Haoyang Zou, Xuefeng Li, Yixiu Liu, Yuxiang Zheng, Ethan Chern, Shijie Xia, Yiwei Qin, Weizhe Yuan, and Pengfei Liu.
\newblock O1 replication journey -- part 2: Surpassing o1-preview through simple distillation, big progress or bitter lesson?, 2024.

\bibitem[JHG{\etalchar{+}}24]{livecodebench}
Naman Jain, King Han, Alex Gu, Wen-Ding Li, Fanjia Yan, Tianjun Zhang, Sida Wang, Armando Solar-Lezama, Koushik Sen, and Ion Stoica.
\newblock Livecodebench: Holistic and contamination free evaluation of large language models for code.
\newblock In {\em The Thirteenth International Conference on Learning Representations}, 2024.

\bibitem[JSA25]{regtext}
Abhinav Java, Simra Shahid, and Chirag Agarwal.
\newblock Towards operationalizing right to data protection.
\newblock In {\em Proceedings of the 2025 Conference of the Nations of the Americas Chapter of the Association for Computational Linguistics: Human Language Technologies (Volume 1: Long Papers)}, pages 8191--8205, 2025.

\bibitem[KGW{\etalchar{+}}23]{redlist}
John Kirchenbauer, Jonas Geiping, Yuxin Wen, Jonathan Katz, Ian Miers, and Tom Goldstein.
\newblock A watermark for large language models.
\newblock In {\em International Conference on Machine Learning}, pages 17061--17084. PMLR, 2023.

\bibitem[Lab25]{bespoke_stratos}
Bespoke Labs.
\newblock Bespoke-stratos: The unreasonable effectiveness of reasoning distillation.
\newblock https://www.bespokelabs.ai/blog/bespoke-stratos-the-unreasonable-effectiveness-of-reasoning-distillation, 2025.
\newblock Accessed: 2025-01-22.

\bibitem[LCG{\etalchar{+}}25]{structure}
Dacheng Li, Shiyi Cao, Tyler Griggs, Shu Liu, Xiangxi Mo, Eric Tang, Sumanth Hegde, Kourosh Hakhamaneshi, Shishir~G. Patil, Matei Zaharia, Joseph~E. Gonzalez, and Ion Stoica.
\newblock Llms can easily learn to reason from demonstrations structure, not content, is what matters!, 2025.

\bibitem[LCX{\etalchar{+}}25]{surveyunlearnabledata}
Jiahao Li, Yiqiang Chen, Yunbing Xing, Yang Gu, and Xiangyuan Lan.
\newblock A survey on unlearnable data, 2025.

\bibitem[LGG{\etalchar{+}}24]{rho}
Zhenghao Lin, Zhibin Gou, Yeyun Gong, Xiao Liu, Yelong Shen, Ruochen Xu, Chen Lin, Yujiu Yang, Jian Jiao, Nan Duan, and Weizhu Chen.
\newblock Not all tokens are what you need for pretraining.
\newblock In A.~Globerson, L.~Mackey, D.~Belgrave, A.~Fan, U.~Paquet, J.~Tomczak, and C.~Zhang, editors, {\em Advances in Neural Information Processing Systems}, volume~37, pages 29029--29063. Curran Associates, Inc., 2024.

\bibitem[LKB{\etalchar{+}}23]{math500}
Hunter Lightman, Vineet Kosaraju, Yuri Burda, Harrison Edwards, Bowen Baker, Teddy Lee, Jan Leike, John Schulman, Ilya Sutskever, and Karl Cobbe.
\newblock Let's verify step by step.
\newblock In {\em The Twelfth International Conference on Learning Representations}, 2023.

\bibitem[LL23]{hint}
Xinzhe Li and Ming Liu.
\newblock Make text unlearnable: Exploiting effective patterns to protect personal data.
\newblock In {\em Proceedings of the 3rd Workshop on Trustworthy Natural Language Processing (TrustNLP 2023)}, pages 249--259, 2023.

\bibitem[LLZM24]{cot}
Zhiyuan Li, Hong Liu, Denny Zhou, and Tengyu Ma.
\newblock Chain of thought empowers transformers to solve inherently serial problems.
\newblock In {\em The Twelfth International Conference on Learning Representations}, 2024.

\bibitem[LTQ{\etalchar{+}}25]{doge}
Pingzhi Li, Zhen Tan, Huaizhi Qu, Huan Liu, and Tianlong Chen.
\newblock Doge: Defensive output generation for llm protection against knowledge distillation.
\newblock {\em arXiv preprint arXiv:2505.19504}, 2025.

\bibitem[LYC{\etalchar{+}}25]{acereason}
Zihan Liu, Zhuolin Yang, Yang Chen, Chankyu Lee, Mohammad Shoeybi, Bryan Catanzaro, and Wei Ping.
\newblock Acereason-nemotron 1.1: Advancing math and code reasoning through sft and rl synergy, 2025.

\bibitem[{MAA}24]{AIME2024}
{MAA}.
\newblock American invitational mathematics examination (aime), 2024.

\bibitem[{Ope}24a]{gpt4o}
{OpenAI}.
\newblock Gpt-4o system card.
\newblock {\em arXiv preprint arXiv:2410.21276}, 2024.

\bibitem[{Ope}24b]{o1}
{OpenAI}.
\newblock Openai o1 system card.
\newblock {\em arXiv preprint arXiv:2412.16720}, 2024.

\bibitem[PLH{\etalchar{+}}25]{distillationwatermark}
Leyi Pan, Aiwei Liu, Shiyu Huang, Yijian Lu, Xuming Hu, Lijie Wen, Irwin King, and Philip~S. Yu.
\newblock Can llm watermarks robustly prevent unauthorized knowledge distillation?, 2025.

\bibitem[QY{\etalchar{+}}25]{qwen25}
Qwen, :, An~Yang, Baosong Yang, Beichen Zhang, Binyuan Hui, Bo~Zheng, Bowen Yu, Chengyuan Li, Dayiheng Liu, Fei Huang, Haoran Wei, Huan Lin, Jian Yang, Jianhong Tu, Jianwei Zhang, Jianxin Yang, Jiaxi Yang, Jingren Zhou, Junyang Lin, Kai Dang, Keming Lu, Keqin Bao, Kexin Yang, Le~Yu, Mei Li, Mingfeng Xue, Pei Zhang, Qin Zhu, Rui Men, Runji Lin, Tianhao Li, Tianyi Tang, Tingyu Xia, Xingzhang Ren, Xuancheng Ren, Yang Fan, Yang Su, Yichang Zhang, Yu~Wan, Yuqiong Liu, Zeyu Cui, Zhenru Zhang, and Zihan Qiu.
\newblock Qwen2.5 technical report, 2025.

\bibitem[RHS{\etalchar{+}}23]{gpqa}
David Rein, Betty~Li Hou, Asa~Cooper Stickland, Jackson Petty, Richard~Yuanzhe Pang, Julien Dirani, Julian Michael, and Samuel~R Bowman.
\newblock Gpqa: A graduate-level google-proof q\&a benchmark.
\newblock In {\em First Conference on Language Modeling}, 2023.

\bibitem[STF{\etalchar{+}}25]{antidistillationsampling}
Yash Savani, Asher Trockman, Zhili Feng, Avi Schwarzschild, Alexander Robey, Marc Finzi, and J~Zico Kolter.
\newblock Antidistillation sampling.
\newblock {\em arXiv preprint arXiv:2504.13146}, 2025.

\bibitem[Tea25]{qwq}
Qwen Team.
\newblock {QwQ-32B}: Embracing the power of reinforcement learning, March 2025.

\bibitem[XLT{\etalchar{+}}24]{kdsurveyllm}
Xiaohan Xu, Ming Li, Chongyang Tao, Tao Shen, Reynold Cheng, Jinyang Li, Can Xu, Dacheng Tao, and Tianyi Zhou.
\newblock A survey on knowledge distillation of large language models, 2024.

\bibitem[YZH{\etalchar{+}}24]{qwenmath}
An~Yang, Beichen Zhang, Binyuan Hui, Bofei Gao, Bowen Yu, Chengpeng Li, Dayiheng Liu, Jianhong Tu, Jingren Zhou, Junyang Lin, Keming Lu, Mingfeng Xue, Runji Lin, Tianyu Liu, Xingzhang Ren, and Zhenru Zhang.
\newblock Qwen2.5-math technical report: Toward mathematical expert model via self-improvement, 2024.

\bibitem[ZLL{\etalchar{+}}25]{qwen3embedding}
Yanzhao Zhang, Mingxin Li, Dingkun Long, Xin Zhang, Huan Lin, Baosong Yang, Pengjun Xie, An~Yang, Dayiheng Liu, Junyang Lin, Fei Huang, and Jingren Zhou.
\newblock Qwen3 embedding: Advancing text embedding and reranking through foundation models.
\newblock {\em arXiv preprint arXiv:2506.05176}, 2025.

\bibitem[ZWP{\etalchar{+}}25]{AM}
Han Zhao, Haotian Wang, Yiping Peng, Sitong Zhao, Xiaoyu Tian, Shuaiting Chen, Yunjie Ji, and Xiangang Li.
\newblock 1.4 million open-source distilled reasoning dataset to empower large language model training, 2025.

\bibitem[ZZZ{\etalchar{+}}24]{llamafactory}
Yaowei Zheng, Richong Zhang, Junhao Zhang, Yanhan Ye, Zheyan Luo, Zhangchi Feng, and Yongqiang Ma.
\newblock Llamafactory: Unified efficient fine-tuning of 100+ language models.
\newblock In {\em Proceedings of the 62nd Annual Meeting of the Association for Computational Linguistics (Volume 3: System Demonstrations)}, Bangkok, Thailand, 2024. Association for Computational Linguistics.

\end{thebibliography}
\bibliographystyle{alpha}

\newpage

\appendix
\section{Related Work}

\paragraph{Reasoning Distillation}

With the success of test-time scaling \cite{o1,R1,qwq}, an increasing number of studies have focused on distilling reasoning ability into smaller models. O1 Journey demonstrates that a base model fine-tuned on only tens of thousands of reasoning traces can outperform O1-preview \cite{o1replicationjourney}. DeepSeek-R1 adopts reinforcement learning for training, followed by distillation to obtain efficient smaller models \cite{R1}. In addition, several datasets have collected large-scale reasoning traces from advanced reasoning models—ranging from tens of thousands to millions—which have been used to train strong distilled models \cite{bespoke_stratos,openthoughts,AM}.

\paragraph{LLM Watermarking}

LLM watermarking focuses on tracking the text generated by LLMs and identifying whether a given piece of text was produced by a particular LLM. Common approaches achieve this by manipulating the sampling distribution during generation, ensuring detectability without compromising readability or fluency \cite{redlist,scalablewatermark}. \cite{distillationwatermark} further explores whether the watermark can still be detected when a student model is distilled using outputs from a protected LLM. While LLM watermarking is also related to model intellectual property detection, its primary emphasis is on post-hoc detection rather than proactively interfering with distillation.

\paragraph{Unlearnable Data}

Unlearnable data focuses on perturbing training data to degrade model performance \cite{surveyunlearnabledata}. \cite{hint} introduces hints into the input text, such as inserting class-wise symbols. RegText treats low-frequency, task-representative tokens as spurious words and randomly inserts these spurious words into the text. These approaches emphasize modifications to the input data, inducing models to rely on shortcuts \cite{regtext}. However, they are unsuitable for antidistillation, since we cannot alter the prompts used by the attacker. Moreover, such methods are typically limited to classification tasks. In contrast, our approach modifies the model’s generation and does not rely on task-specific designs.

\paragraph{Antidistillation}

Recent work has also begun to explore antidistillation for reasoning models. Antidistillation Sampling poisons reasoning traces by modifying a model’s next-token probability distribution during sampling \cite{antidistillationsampling}. This method requires two auxiliary models: a proxy student model, and a variant of the proxy model obtained by performing a single step of gradient ascent on the downstream loss. At each reasoning step, the difference between the logits of these two models is computed to form a perturbation vector. Another method DOGe defends against distillation by fine-tuning the teacher model itself, jointly minimizing the SFT loss while maximizing the KL divergence between the teacher model and the proxy student model \cite{doge}. Both approaches interfere with the teacher model—either by altering its sampling distribution or modifying its parameters. Moreover, their effectiveness has only been demonstrated on small student models ($< $ 4B parameters). By contrast, \ours{} introduces reasoning traces reformulation that do not affect the teacher model’s ability to generate correct answers, and has been validated as effective across student models up to 32B in scale.

\section{Prompts}
\label{sec:prompt}

\begin{tcolorbox}[title=Removing Self-talk Behaviors]
Rewrite the given text, which is a part of a complete reasoning process. Convert only the parts expressed in a self-talk style into a declarative format. Avoid using first-person expressions such as 'I', 'me', 'we', or 'let's'. Do not alter any parts that are not self-talk; keep them exactly as in the original text.

Do not add any extra information. Do not include any introductory phrases.

Text:

\end{tcolorbox}

\begin{tcolorbox}[title=Reordering the Sub-conclusions]
You will process the given text in two steps. The given text is a part of a complete reasoning process.

Step 1: Extract and list the most important sub-conclusions in the given reasoning process. Keep the number of sub-conclusions small and focused.   

Wrap the sub-conclusions in the tags \verb|<SUB>| and \verb|</SUB>| for easy extraction. 

Step 2: Move the sentences corresponding to these sub-conclusions to appear *before* their respective reasoning processes. Keep the sub-conclusions unnumbered and naturally integrated into the context. Do not modify any other parts of the original text.  

Wrap the entire transformed text in the tags \verb|<REWRITTEN>| and \verb|</REWRITTEN>| for easy extraction. 

Text:

\end{tcolorbox}

\section{Hyperparameter}
\label{sec:hyperparameter}

\captionsetup[table]{width=0.7\textwidth}
\begin{table}[ht]
\caption{Hyperparameters used in training. We use larger learning rate for smaller model.}
\vspace{6pt}
\centering
\begin{tabular}{lcc}
\toprule
\textbf{Hyperparameter} & \textbf{Default} & \textbf{1.5B Model} \\
\midrule
Cutoff length   & \multicolumn{2}{c}{16{,}384} \\
Batch size      & \multicolumn{2}{c}{128} \\
Learning rate   & 3.0e-5 & 1.0e-4 \\
Epochs          & \multicolumn{2}{c}{3} \\
LR scheduler    & \multicolumn{2}{c}{Cosine} \\
Warmup ratio    & \multicolumn{2}{c}{0.1} \\
\bottomrule
\end{tabular}
\end{table}

\begin{table}[ht]
\centering
\caption{Hyperparameters used in evaluation. For AIME24, we increase the number of responses per query due to the small number of problems.\label{tab:eval_hyperparams}}
\begin{tabular}{lcc}
\toprule
\textbf{Hyperparameter} & \textbf{Default} & \textbf{AIME24} \\
\midrule
Temperature & \multicolumn{2}{c}{0.6} \\
Top-$p$ & \multicolumn{2}{c}{0.95} \\
Maximum number of tokens to generate & \multicolumn{2}{c}{32{,}768} \\
Number of responses per query & 4 & 16 \\
\bottomrule
\end{tabular}
\end{table}

\section{Example of Reformulated Reasoning Trace}
\label{sec:example}

\noindent
\begin{minipage}[t]{0.48\textwidth}
\begin{tcolorbox}[title=Original Traces, equal height group=A]
Okay, let's see. I need to solve this problem where three people (Chef, head server, sous-chef) have transceivers that can communicate directly if within R meters. If not, but there's a third person acting as a bridge, then they can still communicate. The goal is to check if all three can communicate with each other, possibly through the third.

Hmm. So for each test case, given R and three points, determine if the three form a connected graph where each pair is either directly connected (distance $\le$ R) or connected via the third.

Wait, but how exactly? Let's think. The communication can go through one intermediary. So all three must be in a chain where each consecutive pair is within R, or perhaps any two are connected via a path through the third.
\end{tcolorbox}
\end{minipage}\hfill
\begin{minipage}[t]{0.48\textwidth}
\begin{tcolorbox}[title=Reformulated Traces, equal height group=A]
The goal is to check if all three can communicate with each other, possibly through the third. The problem involves three people (Chef, head server, sous-chef) who have transceivers that can communicate directly if within R meters. If not, but there's a third person acting as a bridge, then they can still communicate.

For each test case, given R and three points, it is necessary to determine if the three form a connected graph where each pair is either directly connected (distance $\le$ R) or connected via the third.

The condition is that all three can communicate with each other, possibly through one another. The communication can go through one intermediary. Therefore, all three must be in a chain where each consecutive pair is within R, or any two are connected via a path through the third.
\end{tcolorbox}
\end{minipage}

\section{Lexical Similarity}
\label{sec:thres}

\begin{tcolorbox}[title=Similarity: 0.63]
\textbf{Original Step: }

But wait, let me verify this again to be sure. Because sometimes with clock angle problems, there can be another instance where the angle is 110 degrees, but depending on the direction (whether the minute hand is ahead or behind the hour hand). Let me think.

\vspace{1em}
\textbf{Matched Part of Reformulated Trace: }

Verification is necessary to ensure accuracy, as there can be another instance where the angle is 110 degrees, depending on the direction of the hands. The formula \textbar 30H - 5.5M\textbar\ provides the absolute angle between the two hands.
\end{tcolorbox}

\begin{tcolorbox}[title=Similarity: 0.75]
\textbf{Original Step: }

Okay, let me try to figure out this problem. So, we have a set of consecutive positive integers starting from 1, and one number is erased. The average of the remaining numbers is 35 and 7/17. We need to find out which number was erased. The options are from 6 to 9, or cannot be determined. Hmm.

\vspace{1em}
\textbf{Matched Part of Reformulated Trace: }

The problem involves a set of consecutive positive integers starting from 1, with one number erased. The average of the remaining numbers is 35 and 7/17. The objective is to identify the erased number. The options range from 6 to 9, or indicate that the number cannot be determined.
\end{tcolorbox}

\begin{tcolorbox}[title=Similarity: 0.81]
\textbf{Original Step: }

Wait, just to be thorough, let's make sure none of the other options could work. Let's check option E: y = 10\^{}x. If we take x = 0, then y = 1, which would correspond to the rotated point (0,1), which is correct. But let's take another point. If x = -1, y = 10\^{}(-1) = 0.1, but according to G', when x = -1, y should be 10. But according to option E, y = 10\^{}(-1) = 0.1, which is wrong. So E is out. Option D, as we saw, gives y = 10\^{}1 = 10 when x = -1, which is correct.

\vspace{1em}
\textbf{Matched Part of Reformulated Trace: }

Check option E: y = 10\^{}x. Taking x = 0, then y = 1, which would correspond to the rotated point (0,1), which is correct. However, taking another point: if x = -1, y = 10\^{}(-1) = 0.1, but according to G', when x = -1, y should be 10. According to option E, y = 10\^{}(-1) = 0.1, which is wrong. Thus, E is out. Option D gives y = 10\^{}1 = 10 when x = -1, which is correct.
\end{tcolorbox}

\begin{tcolorbox}[title=Similarity: 0.96]
\textbf{Original Step: }

So Thuy is 21, Kareem 22. So Kareem is indeed higher than Thuy, and Jose is the lowest of the three. Wait, but Thuy is 21, Jose is 20, Kareem 22. So the order from highest to lowest is Kareem, Thuy, Jose. So the largest is Kareem. Therefore, answer C.

\vspace{1em}
\textbf{Matched Part of Reformulated Trace: }

So Thuy is 21, Kareem 22. So Kareem is indeed higher than Thuy, and Jose is the lowest of the three. Thuy is 21, Jose is 20, Kareem 22. So the order from highest to lowest is Kareem, Thuy, Jose. So the largest is Kareem. Therefore, answer C.
\end{tcolorbox}

\end{document}